# Shannon, Tsallis and Kaniadakis entropies in bi-level image thresholding

**Amelia Carolina Sparavigna**
Department of Applied Science and Technology, Politecnico di Torino, Italy.

**Abstract:** The maximum entropy principle is often used for bi-level or multi-level thresholding of images. For this purpose, some methods are available based on Shannon and Tsallis entropies. In this paper, we discuss them and propose a method based on Kaniadakis entropy.

### 1. Introduction

The concept of entropy was born in thermodynamics and statistical mechanics. Shannon, in 1948, formulated it for the theory of information, obtaining the "information entropy". In an intuitive understanding of it [1], this entropy relates to the amount of uncertainty about an event associated with a given probability distribution.

In image processing, Shannon entropy was the first being used. Today, it is the Tsallis formulation of entropy that seems to be preferred [2-4]. For the elaboration of images, the entropy uses their histograms. For instance, in the bi-level segmentation of a gray-level image, a threshold is determined which separates the gray tones in two systems $A$ and $B$, maximizing the entropy. Considering $A$ and $B$ independent, the entropy $S(A \cup B)$ is the generalized sum $S^A \oplus S^B$, where $S^A, S^B$ are the corresponding entropies of the systems. In this paper, we will discuss the use of Shannon and Tsallis entropies for image thresholding. Among the other formulations of entropy [5], here, we propose the thresholding using Kaniadakis entropy, which is a quite attractive entropy based on the relativistic formulation of the statistical mechanics [6,7].

Both Tsallis and Kaniadakis entropies have an entropic index. If these entropies are used for a bi-level thresholding of an image, the bi-level black and white image that we obtain depends on the value of the optimize threshold, which is depending on the entropic index. Here we compare the results we can obtain with Kaniadakis and Tsallis entropies, proposing a "measure" on the output image. In this "measure" we evaluate the number of edge pixels which separate black and white regions. After experiments on some images, we can conclude that the two entropies compare positively. We have the same results, but Kaniadakis entropy has the intuitive advantage of recovering the Shannon result when its entropic index goes to zero.

### 2. Images and information entropy

Before proposing the use of Kaniadakis entropy for image thresholding, let us shortly illustrate the classic method using the Shannon entropy applied for segmentation. Let us consider and image which as $N$ elements of luminance. These elements can have $g$ gray tones, labelled $\{x_1, x_2, ..., x_n\}$, for instance $\{0,1,2,...,g\}$, with $g = 255$. Let us suppose that each tone $x_i$ is chosen $N_i$ times. The frequency is $f_i = N_i / N$. Accordingly, we have a set $F = \{f_1, f_2, ..., f_n\}$, that we call "the scene". Any given scene has a certain multiplicity $W_F$:

$$W_F = \frac{N!}{(Nf_1)! \ldots (Nf_n)!} \qquad (1)$$

Equation (1) gives the number of ways in which we can generate the same scene [8]. Let us consider that each of the generated copy of the scene has the same probability, which is equal to $1/W_F$. In the case $N$ is large, we can apply the Stirling approximation:

$$\frac{1}{N} \ln W_F = \frac{1}{N} \ln\left(\frac{N!}{(Nf_1)!\ldots(Nf_n)!}\right) = \frac{1}{N}\{\ln N! - [\ln(Nf_1)! + \ldots + \ln(Nf_n)!]\}$$

$$= \frac{1}{N} N \ln N - N - \frac{1}{N}[Nf_1 \ln Nf_1 + Nf_1 + \ldots + Nf_n \ln Nf_n + Nf_n]$$

$$= \frac{1}{N} N \ln N - \frac{1}{N}[Nf_1 \ln N + Nf_1 \ln f_1 + \ldots + Nf_n \ln N + Nf_n \ln f_n] \quad (2)$$

$$= -[f_1 \ln f_1 + \ldots + f_n \ln f_n] = -\sum_{i=1}^{n} f_i \ln f_i$$

In (2), we use $\sum_{i=1}^{n} f_i = 1$. This is the Shannon entropy; it is depending on "objective" frequencies $f_i$ instead of "subjective" probabilities $p_i$ [8].

### 3. Thresholding

For a bi-level thresholding of an image, let us follow the approach of [4].

Let us consider two independent systems $A$ and $B$, for which the joint probability is $p(A,B) = p(A)p(B)$. The entropy $S(A \cup B)$ is $S^A \oplus S^B$. The systems can be given as in the following [4]. $A$ contains elements with $g$ gray tones, labelled $\{x_1, x_2, \ldots, x_t\}$, for instance $\{0,1,2,\ldots,t\}$, that is, with gray tone below or equal a given threshold $t$. Let us suppose that each tone $x_i$ is chosen $N_{A,i}$ times, according to frequency: $f_{A,i} = f_i N / N_A$. We have: $N_A = \sum_{i=1}^{t} Nf_i = N \sum_{i=1}^{t} f_i = NP_A$.

Therefore: $f_{A,i} = Nf_i /(NP_A) = f_i / P_A$. Moreover, $\sum_{i=1}^{t} f_{A,i} = 1$. In the same manner, $B$ contains elements $\{x_{t+1}, x_2, \ldots, x_n\}$, that is, $\{t+1, t+2, \ldots, g\}$. Let us suppose that each tone $x_i$ is chosen $N_{B,i}$ times, according to frequency: $f_{B,i} = f_i N / N_B$. Again: $N_B = \sum_{i=t+1}^{g} Nf_i = N \sum_{i=t+1}^{g} f_i = NP_B$.

The entropy $S^A$ is:

$$S^A = \frac{1}{N_A} \ln W_A = \frac{1}{N_A} \ln\left(\frac{N_A!}{(N_{A,1})!\ldots(N_{A,t})!}\right) \quad (3)$$

Then, using Stirling again:

$$S^A = -\sum_{i=1}^{t} f_{A,i} \ln f_{A,i} = -\sum_{i=1}^{t} \frac{f_i}{P_A} \ln \frac{f_i}{P_A} \quad (4)$$

For B:

$$S^B = \frac{1}{N_B} \ln W_B = \frac{1}{N_B} \ln\left(\frac{N_B!}{(N_{B,t+1})!\ldots(N_{B,g})!}\right) = -\sum_{i=t+1}^{g} \frac{f_i}{P_B} \ln \frac{f_i}{P_B} \quad (5)$$

The abovementioned frequencies for a gray-level image can be given by the normalized histogram. To find the best value of threshold t, we have to maximize $S^A + S^B$ [4].

**3. Bi-level thresholding with Tsallis and Kaniadakis entropies**

Let us remember that the Tsallis entropy is given by [9,10]:

$$S_q = \frac{1}{q-1}\left\{1 - \sum_i p_i^q\right\} = \frac{1}{q-1}\left\{\sum_i p_i\left(1 - p_i^{q-1}\right)\right\} \qquad (6)$$

In fact, the Tsallis entropy is defined, using the $q$-logarithm, as:

$$S_q = -\sum_i p_i^q \ln_q p_i = -\sum_i p_i^q \frac{p_i^{1-q} - 1}{1-q} = -\sum_i \frac{p_i - p_i^q}{1-q} = \frac{1}{q-1}\left\{1 - \sum_i p_i^q\right\} \qquad (7)$$

Let us assume a bi-level threshold $t$ for the gray levels. In [3], two systems had been introduced, $A$ and $B$, and their probability distributions. Let us assume the properties of $A$ and $B$ as in the previous section. The Tsallis entropies, one for each distribution, are given by:

$$S_q^A(t) = \frac{1}{q-1}\left\{1 - \sum_{i=1}^{t}\left(\frac{f_i}{P_A}\right)^q\right\} \qquad (8)$$

$$S_q^B(t) = \frac{1}{q-1}\left\{1 - \sum_{i=t+1}^{g}\left(\frac{f_i}{P_B}\right)^q\right\} \qquad (9)$$

Taking the limit $q \to 1$, Tsallis entropy gives Shannon's entropy. The total Tsallis entropy is given by the generalized sum:

$$S_q(t) = S_q^A(t) + S_q^B(t) + (1-q)S_q^A(t)S_q^B(t) \qquad (10)$$

In fact, in a generalization of statistical mechanics, a deformed entropy had been proposed, the Kaniadakis entropy, also known as κ-entropy [6,7]:

$$S_\kappa = -\sum_i p_i \ln_{\{\kappa\}}(p_i) \qquad (11)$$

This entropy has the remarkable property of having the same behavior of Shannon entropy, that is:

$$S_\kappa = \sum_i p_i \ln_{\{\kappa\}}\left(\frac{1}{p_i}\right) \qquad (12)$$

In it, we have the generalized version of the logarithm [7]:

$$S_\kappa = -\frac{1}{2\kappa}\left\{\sum_i \left((p_i)^{1+k} - (p_i)^{1-k}\right)\right\} \qquad (13)$$

We can apply this entropy to the bi-level thresholding. Let us call the threshold τ.
Generalizing (3) and (5), κ-entropies are:

$$S^A_\kappa(\tau) = -\frac{1}{2\kappa}\left\{\sum_{i=1}^{\tau}\left[\left(\frac{f_i}{P_A}\right)^{1+k} - \left(\frac{f_i}{P_A}\right)^{1-k}\right]\right\} \quad (14)$$

$$S^B_\kappa(\tau) = -\frac{1}{2\kappa}\left\{\sum_{i=\tau+1}^{g}\left[\left(\frac{f_i}{P_B}\right)^{1+k} - \left(\frac{f_i}{P_B}\right)^{1-k}\right]\right\} \quad (15)$$

In the limit $\kappa \to 0$, Kaniadakis entropy becomes Shannon entropy.
Let us consider the composition of systems *A* and *B*, but in the framework of this deformed statistics. According to [11], the generalized sum:

$$S_\kappa(\tau) = S^A_\kappa(\tau)\Im^B_\kappa(\tau) + S^B_\kappa(\tau)\Im^A_\kappa(\tau) \quad (16)$$

In (14) we have:

$$\Im^A_\kappa(\tau) = \frac{1}{2}\left\{\sum_{i=1}^{\tau}\left[\left(\frac{f_i}{P_A}\right)^{1+k} + \left(\frac{f_i}{P_A}\right)^{1-k}\right]\right\} \quad (17)$$

$$\Im^B_\kappa(\tau) = \frac{1}{2}\left\{\sum_{i=\tau+1}^{g}\left[\left(\frac{f_i}{P_B}\right)^{1+k} + \left(\frac{f_i}{P_B}\right)^{1-k}\right]\right\} \quad (18)$$

When entropies (10) or (16) are maximized, the corresponding gray level τ is considered the optimum threshold value. In the gray bi-level thresholding, we have a resulting processed image, which is a black and white image. The output image is created as in the following: if pixels have a gray tone larger than the threshold, they become white. If pixels have a lower value, they become black.

**4. More deeply in the limit of Kaniadakis entropy**
In fact, we have that:

$$\Im_\kappa = \frac{1}{2}\left\{\sum_i p_i^{1+\kappa} + \sum_i p_i^{1-\kappa}\right\} = \kappa S_\kappa + \sum_i p_i^{1+\kappa} \quad (19)$$

In fact:

$$\Im_\kappa = \frac{1}{2}\left\{\sum_i p_i^{1+\kappa} + \sum_i p_i^{1-\kappa}\right\} = -\kappa\frac{1}{2\kappa}\left\{\sum_i p_i^{1+\kappa} - \sum_i p_i^{1-\kappa}\right\} + \sum_i p_i^{1+\kappa}$$

$$= -\frac{1}{2}\sum_i p_i^{1+\kappa} + \frac{1}{2}\sum_i p_i^{1-\kappa} + \sum_i p_i^{1+\kappa} = \Im_\kappa$$

Therefore:

$$S_\kappa = S^A_\kappa \Im^B_\kappa + S^B_\kappa \Im^A_\kappa$$

$$= S^A_\kappa\left(\kappa S^B_\kappa + \sum_{i=\tau+1}^{g}\left(\frac{f_i}{P_B}\right)^{1+\kappa}\right) + S^B_\kappa\left(\kappa S^A_\kappa + \sum_{i=1}^{\tau}\left(\frac{f_i}{P_A}\right)^{1+\kappa}\right) \quad (20)$$

$$S_\kappa = 2\kappa S^A_\kappa S^B_\kappa + S^A_\kappa \sum_{i=\tau+1}^{g}\left(\frac{f_i}{P_B}\right)^{1+\kappa} + S^B_\kappa \sum_{i=1}^{\tau}\left(\frac{f_i}{P_A}\right)^{1+\kappa} \quad (21)$$

In the limit $\kappa \to 0$, Kaniadakis entropy becomes Shannon entropy, and therefore we must have the normal additivity. And in fact:

$$S_\kappa = S^A \sum_{i=\tau+1}^{g} \frac{f_i}{P_B} + S^B \sum_{i=1}^{\tau} \frac{f_i}{P_A} = S^A + S^B \qquad (22)$$

$S^A, S^B$ are the Shannon entropies. In the limit $\sum_{i=1}^{\tau}\left(\frac{f_i}{P_A}\right)^{1+\kappa}$ becomes $\sum_{i=1}^{\tau}\frac{f_i}{P_A} = \frac{1}{P_A}\sum_{i=1}^{\tau} f_i = \frac{P_A}{P_A} = 1$. We have the same for $B$.

### 4. Discussion

Let us note that, both Tsallis and Kaniadakis entropies have entropic indices that can give different results when applied to the sample. To choose among these several results and define an output image, we propose a "measure" of the bi-level image, given by the number of edge pixels between black and white regions. Of course, other measures can be defined.

Figures 1-3 give images and bi-level images. The corresponding Tables are the results of maximizing Tsallis and Kaniadakis entropies, according to (10) and (16). In experiments, their entropic indices are spanning interval (0,1). Let us avoid, in the calculations, the values 0 and 1. We can see that the two entropies are able to provide the same results. We have also, in the case of Figure 3 and Table VI, the evidence of an "image transition" (see Ref.12 for more details).

Besides the fact that the Kaniadakis entropy possesses a formalism which is closer to that of Shannon entropy, a good reason for preferring κ-entropy is that it has the more intuitive behavior of an entropy recovering the Shannon one, when its entropic index is going to zero. Another relevant advantage of Kaniadakis entropy is in the evaluation of multi-level thresholding of images. This will be discussed in a future paper.

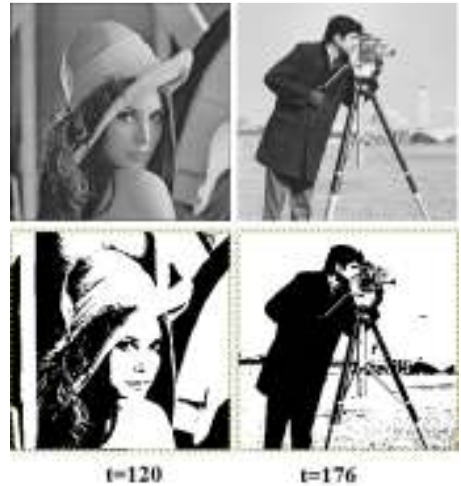

Fig.1. Lena and Cameraman and the corresponding bi-level images, *t* is the threshold value.

| Entropic indices (q,κ) | Threshold (T-entropy) | Number of edge pixels | Threshold (κ-entropy) | Number of edge pixels |
|---|---|---|---|---|
| 0.01 | 119 | 7807 | 120 | 7901 |
| 0.05 | 119 | 7807 | 120 | 7901 |
| 0.1 | 119 | 7901 | 120 | 7901 |
| 0.15 | 119 | 7901 | 120 | 7901 |
| 0.2 | 120 | 7901 | 120 | 7901 |
| 0.4 | 120 | 7901 | 120 | 7901 |
| 0.6 | 120 | 7901 | 120 | 7901 |
| 0.8 | 120 | 7901 | 120 | 7901 |
| 0.9 | 120 | 7901 | 119 | 7807 |
| 0.99 | 120 | 7901 | 119 | 7807 |

Table I: Optimized thresholds on Lena, using Tsallis and Kaniadakis entropies, for several values of entropic indices. In the limit $q \to 1$, Tsallis entropy provides Shannon result, and for $\kappa \to 0$, Kaniadakis entropy becomes Shannon entropy. When the image is segmented in a bi-level black and white image according to the given threshold, the number of edge pixels between black and white regions are calculated. If we assume that the "best" bi-level image is that having the largest number of edge pixels, the threshold to choose is 120 from Tsallis and κ-entropy.

| Entropic indices (κ,q) | Threshold (T-entropy) | Number of edge pixels | Threshold (κ-entropy) | Number of edge pixels |
|---|---|---|---|---|
| 0.01 | 138 | 4559 | 176 | 7186 |
| 0.05 | 140 | 4557 | 175 | 6938 |
| 0.1 | 141 | 4579 | 175 | 6938 |
| 0.15 | 143 | 4596 | 174 | 6734 |
| 0.2 | 145 | 4570 | 171 | 6176 |
| 0.3 | 148 | 4628 | 168 | 5716 |
| 0.4 | 154 | 4695 | 165 | 5323 |
| 0.5 | 160 | 4895 | 160 | 4895 |
| 0.6 | 165 | 5323 | 154 | 4695 |
| 0.7 | 167 | 5570 | 148 | 4628 |
| 0.8 | 171 | 6176 | 145 | 4570 |
| 0.9 | 175 | 6938 | 141 | 4579 |
| 0.99 | 175 | 6938 | 138 | 4559 |

Table II: Optimized thresholds for Cameraman. The "best" bi-level image is that having the largest number of edge pixels and the threshold to choose is 175 from Tsallis entropy and 176 from κ-entropy.

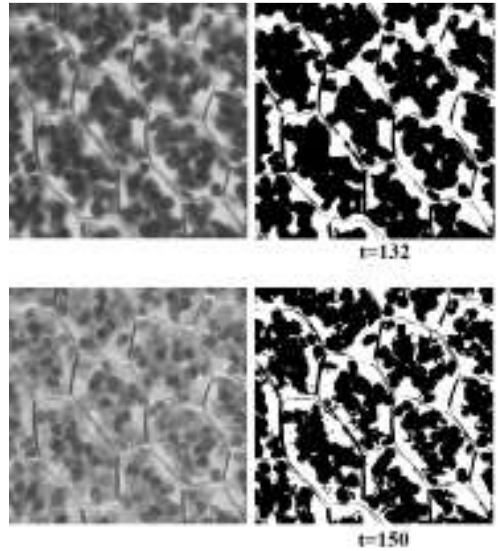

Fig.2. Microscopic image of cells (courtesy Kristian Peters, Wikipedia) in two different gray-level renderings and the corresponding bi-level images.

| Entropic indices (κ,q) | Threshold (T-entropy) | Number of edge pixels | Threshold (κ-entropy) | Number of edge pixels |
|---|---|---|---|---|
| 0.01 | 148 | 14462 | 132 | 15376 |
| 0.05 | 146 | 14596 | 132 | 15376 |
| 0.1 | 145 | 14725 | 132 | 15376 |
| 0.15 | 143 | 14907 | 132 | 15375 |
| 0.2 | 141 | 15000 | 132 | 15376 |
| 0.3 | 139 | 15153 | 135 | 15293 |
| 0.4 | 137 | 15213 | 135 | 15293 |
| 0.5 | 136 | 15229 | 136 | 15229 |
| 0.6 | 135 | 15293 | 137 | 15213 |
| 0.7 | 135 | 15293 | 139 | 15153 |
| 0.8 | 132 | 15376 | 141 | 15000 |
| 0.9 | 132 | 15376 | 145 | 14725 |
| 0.99 | 132 | 15376 | 148 | 14462 |

Table III: Optimized thresholds for the upper image of Figure 2. The "best" bi-level image is that having the largest number of edge pixels and the threshold to choose is 132 from Tsallis and κ-entropy.

| Entropic indices (κ,q) | Threshold (T-entropy) | Number of edge pixels | Threshold (κ-entropy) | Number of edge pixels |
|---|---|---|---|---|
| 0.01 | 154 | 20663 | 150 | 21461 |
| 0.05 | 154 | 20663 | 150 | 21461 |
| 0.1 | 155 | 20450 | 150 | 21461 |
| 0.15 | 155 | 20450 | 151 | 21277 |
| 0.2 | 154 | 20663 | 151 | 21277 |
| 0.3 | 154 | 20663 | 151 | 21277 |
| 0.4 | 154 | 20663 | 153 | 20903 |
| 0.5 | 153 | 20903 | 153 | 20903 |
| 0.6 | 153 | 20903 | 154 | 20663 |
| 0.7 | 151 | 21277 | 154 | 20663 |
| 0.8 | 151 | 21277 | 154 | 20663 |
| 0.9 | 151 | 21277 | 155 | 20450 |
| 0.99 | 150 | 21461 | 154 | 20663 |

Table IV: Optimized thresholds for the lower image of Figure 2. The "best" bi-level image is that having the largest number of edge pixels and the threshold to choose is 150 from Tsallis and κ-entropy.

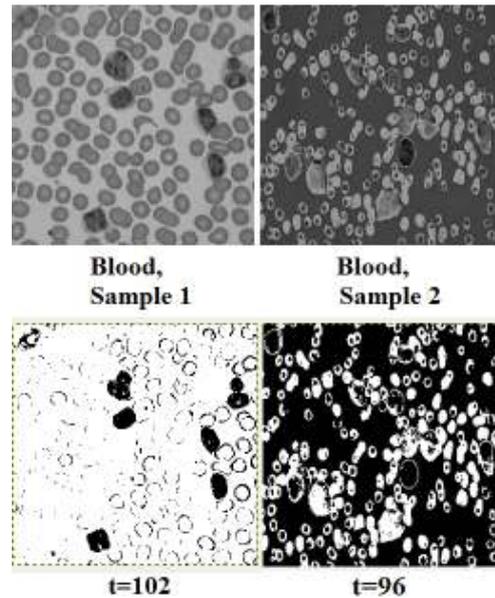

Fig.3. Microscopic image of blood cells (courtesy Wikipedia) and the corresponding bi-level images.

| Entropic indices (κ,q) | Threshold (T-entropy) | Number of edge pixels | Threshold (κ-entropy) | Number of edge pixels |
|---|---|---|---|---|
| 0.01 | 102 | 7215 | 93 | 1525 |
| 0.05 | 98 | 3253 | 93 | 1525 |
| 0.1 | 96 | 2281 | 94 | 1735 |
| 0.2 | 95 | 1966 | 94 | 1735 |
| 0.3 | 95 | 1966 | 95 | 1966 |
| 0.5 | 95 | 1966 | 95 | 1966 |
| 0.7 | 95 | 1966 | 95 | 1966 |
| 0.8 | 94 | 1735 | 95 | 1966 |
| 0.9 | 94 | 1735 | 96 | 2281 |
| 0.99 | 93 | 1525 | 102 | 7215 |

Table V: Optimized thresholds (blood, Sample 1). If we assume, that the "best" bi-level image is that having the largest number of pixels at edges, the threshold to choose it 102.

| Entropic indices (κ,q) | Threshold (T-entropy) | Number of pixels at edges | Threshold (κ-entropy) | Number of pixels at edges |
|---|---|---|---|---|
| 0.01 | 90 | 13561 | 49 | 858 |
| 0.05 | 90 | 13561 | 49 | 858 |
| 0.1 | 91 | 13603 | 49 | 858 |
| 0.2 | 91 | 13603 | 49 | 858 |
| 0.3 | 96 | 13691 | 49 | 858 |
| 0.4 | 96 | 13691 | 49 | 858 |
| 0.5 | 87 | 13530 | 87 | 13530 |
| 0.55 | 49 | 858 | 87 | 13530 |
| 0.6 | 49 | 858 | 96 | 13691 |
| 0.7 | 49 | 858 | 96 | 13691 |
| 0.8 | 49 | 858 | 91 | 13603 |
| 0.9 | 49 | 858 | 91 | 13603 |
| 0.99 | 49 | 858 | 90 | 13561 |

Table VI: Optimized thresholds (blood, Sample 2). Again, the threshold to choose is 96. Note the presence of a texture transition [12].